\documentclass[11pt, a4paper, logo]{deepmind}

\usepackage{hyperref}
\usepackage{url}
\usepackage{booktabs}       
\usepackage{amsfonts}       
\usepackage{nicefrac}       
\usepackage{microtype}      
\usepackage{graphicx}
\usepackage{amsmath}
\usepackage{amssymb}
\usepackage{amsthm}
\usepackage{algorithm}
\usepackage{algorithmic}
\usepackage{color}
\usepackage{subcaption}
\usepackage{sgame, tikz}
\usepackage{listings}
\usepackage{bm}
\usepackage{booktabs}
\usepackage{tikz}
\usepackage{soul}
\usepackage{multirow}
\usepackage[algo2e, noend, noline, ruled, linesnumbered]{algorithm2e}

\providecommand{\DontPrintSemicolon}{\dontprintsemicolon}
\DontPrintSemicolon
\usepackage[utf8]{inputenc}

\title{OpenSpiel: A Framework for Reinforcement Learning in Games}

\author[1,**]{Marc Lanctot}
\author[1,**]{Edward Lockhart}
\author[1,**]{Jean-Baptiste Lespiau}
\author[1,**]{Vinicius Zambaldi}
\author[2]{Satyaki Upadhyay}
\author[1]{Julien P\'{e}rolat}
\author[2]{Sriram Srinivasan}
\author[1]{Finbarr Timbers}
\author[1]{Karl Tuyls}
\author[1]{Shayegan Omidshafiei}
\author[1]{Daniel Hennes}
\author[1,3]{Dustin Morrill}
\author[1]{Paul Muller}
\author[1]{Timo Ewalds}
\author[1]{Ryan Faulkner}
\author[1]{J\'{a}nos Kram\'{a}r}
\author[1]{Bart De Vylder}
\author[2]{Brennan Saeta}
\author[2]{James Bradbury}
\author[1]{David Ding}
\author[1]{Sebastian Borgeaud}
\author[1]{Matthew Lai}
\author[1]{Julian Schrittwieser}
\author[1]{Thomas Anthony}
\author[1]{Edward Hughes}
\author[1]{Ivo Danihelka}
\author[2]{Jonah Ryan-Davis}  

\affil[1]{DeepMind}
\affil[2]{Google}
\affil[3]{University of Alberta}
\affil[**]{These authors contributed equally}

\definecolor{darkgreen}{RGB}{0,125,0}
\definecolor{darkblue}{RGB}{0,0,125}
\definecolor{kellygreen}{RGB}{76, 187, 23}
\definecolor{amber}{RGB}{255, 191, 0}

\newcounter{mlNoteCounter}
\newcounter{vzNoteCounter}
\newcounter{jblNoteCounter}

\newcounter{dmNoteCounter}

\newcommand{\argmax}{\operatornamewithlimits{argmax}}
\newcommand{\bE}{\mathbb{E}}

\newcommand{\cA}{\mathcal{A}}

\newcommand{\cH}{\mathcal{H}}

\newcommand{\cN}{\mathcal{N}}

\newcommand{\cS}{\mathcal{S}}
\newcommand{\cT}{\mathcal{T}}
\newcommand{\cZ}{\mathcal{Z}}

\newcommand{\defword}[1]{\textbf{\boldmath{#1}}}

\newtheorem{definition}{Definition}
\newtheorem{fact}{Fact}

\definecolor{listinggray}{gray}{0.9}
\definecolor{gcolor}{rgb}{0.95,0.95,0.95}
\definecolor{wcolor}{rgb}{1.0,1.0,1.0}
\definecolor{bcolor}{rgb}{0.0,0.0,0.0}
\begin{abstract}
OpenSpiel is a collection of environments and algorithms for research in general reinforcement learning
and search/planning in games.
OpenSpiel supports $n$-player (single- and multi- agent) zero-sum, cooperative and general-sum, one-shot and sequential, 
strictly turn-taking and simultaneous-move, 
perfect and imperfect information games, as well as traditional multiagent environments such as (partially- and fully- observable)
grid worlds and social dilemmas.
OpenSpiel also includes tools to analyze learning dynamics and other common evaluation metrics.
This document serves both as an overview of the code base and an introduction to the terminology, core concepts, and
algorithms across the fields of reinforcement learning, computational game theory, and search.
\end{abstract}

\begin{document}

\maketitle

\vspace{1.5cm}
\begin{center}
\includegraphics[scale=0.24]{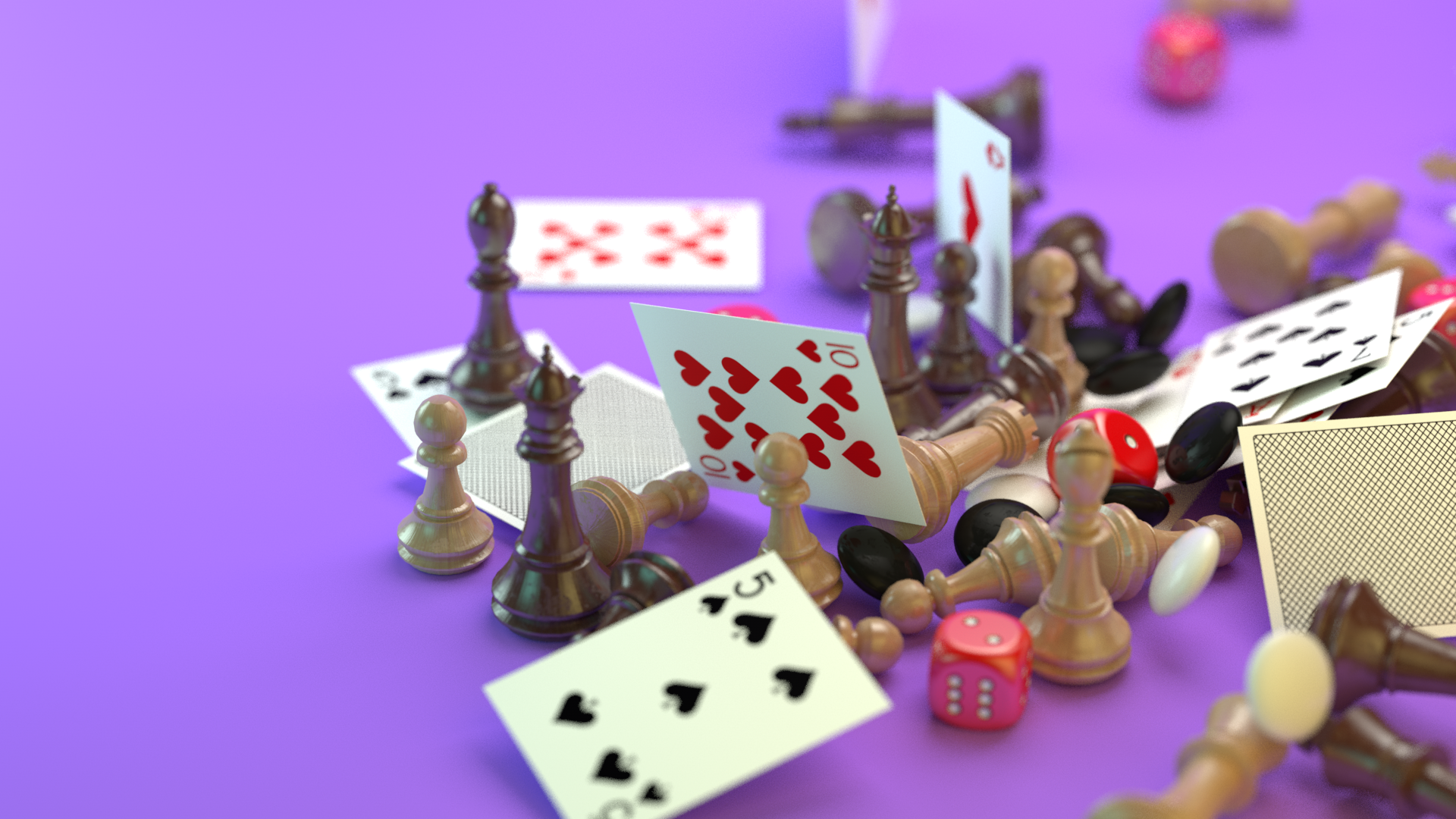}
\end{center}

\newpage

\tableofcontents

\newpage

\section{OpenSpiel Overview}

\subsection{Acknowledgments}

OpenSpiel has been possible due to a team of contributors. For a full list of all the contributors, please see \href{https://github.com/deepmind/open_spiel/blob/master/docs/authors.md}{the list of authors on github}.

We would also like to thank the following people, who helped and supported the development of OpenSpiel:
\begin{itemize}
    \item Remi Munos
    \item Michael Bowling
    \item Thore Graepel
    \item Shibl Mourad
    \item Nathalie Beauguerlange
    \item Ellen Clancy
    \item Louise Deason
    \item Andreas Fidjeland
    \item Michelle Bawn
    \item Yoram Bachrach
    \item Dan Zheng
    \item Martin Schmid
    \item Neil Burch
    \item Damien Boudot
    \item Adam Cain
\end{itemize}

\subsection{OpenSpiel At a Glance}

We provide an intentionally brief overview here. For details, please see Section~\ref{sec:design-api}.

OpenSpiel provides a framework for writing games and algorithms and evaluating them on a variety of benchmark games. 
OpenSpiel contains implementations of over 20 different games of various sorts (perfect information, simultaneous move, imperfect information, gridworld games, an auction game, and several normal-form / matrix games).
Game implementations are in C\texttt{++} and wrapped in Python. Algorithms are implemented in C\texttt{++} and/or Python. The API is almost identical in the two languages, so code can easily be translated if needed. 
A subset of the library has also been ported to Swift.
Most of the learning algorithms written in Python use Tensorflow~\cite{Tensorflow}, though we are actively seeking examples and other support for PyTorch~\cite{Paszke2017automatic} and JAX\footnote{https://github.com/google/jax}.

OpenSpiel has been tested on Linux and MacOS. There is also limited support on Windows.

Components of OpenSpiel are listed in Tables \ref{tab:games} and \ref{tab:algorithms}. As of October 2019, these tables will no
longer be updated: Please refer to the \href{https://github.com/deepmind/open_spiel/blob/master/docs/games.md}{Overview of Implemented Games} or the \href{https://github.com/deepmind/open_spiel/blob/master/docs/algorithms.md}{Overview of Implemented Algorithms} pages
on the web site for most current information.

\vspace{3cm}
\begin{table}[h!]
    \centering
    \begin{tabular}{l|c}
         Game & Reference(s) \\
         \hline
         Backgammon    & \href{https://en.wikipedia.org/wiki/Backgammon}{Wikipedia}   \\
         Breakthrough  & \href{https://en.wikipedia.org/wiki/Breakthrough_(board_game)}{Wikipedia} \\
         Bridge bidding & \href{https://en.wikipedia.org/wiki/Contract_bridge}{Wikipedia} \\
         Catch          & \cite{Mnih14Recurrent} and \cite[Appendix A]{Osband19BSuite} \\
         Coin Game   & \cite{Raileanu18Modeling} \\
         Connect Four & \href{https://en.wikipedia.org/wiki/Connect_Four}{Wikipedia} \\
         Cooperative Box-Pushing & \cite{Seuken12Improved} \\
         Chess       & \href{https://en.wikipedia.org/wiki/Chess}{Wikipedia} \\
         First-price Sealed-bid Auction & \href{https://en.wikipedia.org/wiki/First-price_sealed-bid_auction}{Wikipedia} \\
         Go          & \href{https://en.wikipedia.org/wiki/Go_(game)}{Wikipedia} \\
         Goofspiel   & \href{https://en.wikipedia.org/wiki/Goofspiel}{Wikipedia}  \\
         Hanabi (via \href{https://github.com/deepmind/hanabi-learning-environment}{HLE})   & \href{https://en.wikipedia.org/wiki/Hanabi_(card_game)}{Wikipedia}, \cite{Bard19Hanabi} \\
         Havannah    & \href{https://en.wikipedia.org/wiki/Havannah}{Wikipedia}    \\
         Hex         & \href{https://en.wikipedia.org/wiki/Hex_(board_game)}{Wikipedia}, \cite{HT19}  \\
         Kuhn poker  & \href{https://en.wikipedia.org/wiki/Kuhn_poker}{Wikipedia}, \cite{Kuhn50} \\
         Laser Tag   & \cite{Leibo17SSD,Lanctot17PSRO} \\
         Leduc poker & \cite{Southey05bayes} \\
         Liar's Dice & \href{https://en.wikipedia.org/wiki/Liar\%27s_dice}{Wikipedia} \\
         Markov Soccer & \cite{Littman94markovgames, He16DRON} \\
         Matching Pennies (three-player) & \cite{Jordan93mp3p} \\
         Matrix Games & \cite{Shoham09} \\
         Negotiation & \cite{Lewis17,Cao18Emergent} \\
         Oshi-Zumo & \cite{Buro04OshiZumo,Bosansky16Algorithms,Perolat16Softened} \\
         Oware & \href{https://en.wikipedia.org/wiki/Oware)}{Wikipedia} \\
         Pentago & \href{https://en.wikipedia.org/wiki/Pentago}{Wikipedia}  \\
         Phantom Tic-Tac-Toe & \cite{Auger11Multiple,lanctot13phdthesis,Lisy14selection} \\
         Pig & \cite{Neller04Pig} \\
         Quoridor & \href{https://en.wikipedia.org/wiki/Quoridor}{Wikipedia} \\
         Tic-Tac-Toe & \href{https://en.wikipedia.org/wiki/Tic-tac-toe}{Wikipedia} \\
         Tiny Bridge & \\
         Tiny Hanabi & \cite{Foerster18BAD} \\
         Y & \href{https://en.wikipedia.org/wiki/Y_(game)}{Wikipedia} \\
         \hline
         Cliff-Walking (Python-only) & \cite[Chapter 6]{Sutton18} \\
         \hline
    \end{tabular}
    \caption{Game Implementations in OpenSpiel as of October 2019. Please see \href{https://github.com/deepmind/open_spiel/blob/master/docs/games.md}{Overview of Implemented Games} for an up-to-date list.}
    \label{tab:games}
\end{table}

\begin{table}[h!]
    \centering
    \begin{tabular}{l|c|c}
         Algorithm &  Category & Reference(s) \\
         \hline
         \hline
         Minimax (and Alpha-Beta) Search      & Search &
         \href{https://en.wikipedia.org/wiki/Minimax#Minimax_algorithm_with_alternate_moves}{Wikipedia},
         \href{https://en.wikipedia.org/wiki/Alpha\%E2\%80\%93beta_pruning}{Wikipedia}, \cite{KnuthMoore75} \\
         Monte Carlo tree search & Search & \href{https://en.wikipedia.org/wiki/Monte_Carlo_tree_search}{Wikipedia}, \cite{UCT,Coulom06, mctssurvey} \\
         \hline
         Lemke-Howson (via {\tt nashpy}) & Opt. & \cite{Shoham09} \\
         Sequence-form linear programming & Opt. & \cite{SequenceFormLPs,Shoham09} \\
         \hline
         Counterfactual Regret Minimization (CFR) & Tabular & \cite{CFR,Neller13cfrnotes} \\
         CFR against a best responder (CFR-BR) & Tabular & \cite{Johanson12CFRBR} \\
         Exploitability / Best Response & Tabular & \cite{CFR} \\
         External sampling Monte Carlo CFR & Tabular & \cite{Lanctot09Sampling,lanctot13phdthesis} \\
         Outcome sampling Monte Carlo CFR & Tabular & \cite{Lanctot09Sampling,lanctot13phdthesis} \\
         Q-learning                 & Tabular & \cite{Sutton18}  \\
         Value Iteration            & Tabular & \cite{Sutton18}  \\
        \hline
         Advantage Actor-Critic (A2C) & RL & \cite{Mnih2016asynchronous} \\
         Deep Q-networks (DQN) & RL & \cite{Mnih15DQN}  \\
         Ephemeral Value Adjustments (EVA) & RL & \cite{Hansen18EVA} \\
         \hline
         Deep CFR & MARL & \cite{Brown19DeepCFR}  \\
         Exploitability Descent (ED) & MARL & \cite{Lockhart19ED}  \\
         (Extensive-form) Fictitious Play (XFP) & MARL & \cite{Heinrich15FSP} \\
         Neural Fictitious Self-Play (NFSP) & MARL & \cite{Heinrich16} \\
         Neural Replicator Dynamics (NeuRD) & MARL & \cite{Omidshafiei19NeuRD} \\
         Regret Policy Gradients (RPG, RMPG) & MARL & \cite{Srinivasan18RPG} \\
         Policy-Space Response Oracles (PSRO) & MARL & \cite{Lanctot17PSRO} \\
         Q-based ``all-action'' Policy Gradients (QPG) & MARL & \cite{Sutton01Comparing,Peters02Policy,Srinivasan18RPG} \\
         Regression CFR (RCFR) & MARL & \cite{Waugh15solving,Morrill16}  \\
         Rectified Nash Response ($\textrm{PSRO}_\textrm{rN}$) & MARL & \cite{Balduzzi19}  \\
         \hline
         $\alpha$-Rank & Eval / Viz &  \cite{Omidshafiei19AlphaRank} \\
         Replicator / Evolutionary Dynamics & Eval / Viz &  \cite{HofbauerSigmund98,Sandholm10Population} \\
    \end{tabular}
    \caption{Algorithms Implementated in OpenSpiel as of October 2019. Please see \href{https://github.com/deepmind/open_spiel/blob/master/docs/algorithms.md}{Overview of Implemented Algorithms} for an updated list.}
    \label{tab:algorithms}
\end{table}

\clearpage

\section{Getting Started}

\subsection{Getting and Building OpenSpiel}
\label{sec:building}

The following commands will clone the repository and build OpenSpiel on Ubuntu or Debian Linux, or MacOS.
There is also limited support for Windows. We now show the {\it fastest}
way to install OpenSpiel. Please see the recommended  \href{https://github.com/deepmind/open_spiel/blob/master/docs/install.md}{installation instructions} using {\tt virtualenv} for more detail.

\begin{lstlisting}[language=bash]
apt-get install git
git clone https://github.com/deepmind/open_spiel.git
cd open_spiel
./install.sh
curl https://bootstrap.pypa.io/get-pip.py -o get-pip.py
# Install pip deps as your user. Do not use the system's pip.
python3 get-pip.py --user
# If using Linux:
  export PATH="$PATH:$HOME/.local/bin"
  ln -s $HOME/.local/bin/pip3 $HOME/.local/bin/mypip3
# Else (MacOS):
  export PATH="$PATH:$HOME/Library/Python/3.7/bin"
  ln -s $HOME/Library/Python/3.7/bin/pip3 $HOME/Library/Python/3.7/bin/mypip3
mypip3 install --upgrade pip --user
mypip3 install --upgrade setuptools testresources --user
mypip3 install --upgrade -r requirements.txt --user
mkdir build
cd build
CXX=clang++
cmake -DPython_TARGET_VERSION=3.6 -DCMAKE_CXX_COMPILER=clang++ ../open_spiel
make -j12   # The 12 here is the number of parallel processes used to build 
ctest -j12  # Run the tests to verify that the installation succeeded
\end{lstlisting}

Note that we have tested OpenSpiel Linux and MacOS, and there is limited support on Windows. Also,
for the case of Linux, some of the scripts and instructions currently assume Debian-based distributions
(i.e. Debian, Ubuntu, etc.). All of the dependencies exist on other distributions, but may have different
names, and package managers differ. Please see {\tt install.sh} for necessary dependencies.

\subsubsection{Setting PATH and PYTHONPATH}
\label{sec:pythonpath}

To be able to import the Python code (both the C\texttt{++} binding pyspiel and the rest) from any location, you will need to add to your
PYTHONPATH the root directory and the \texttt{open\_spiel} directory. Add the following in your \texttt{.bashrc} or \texttt{.profile}:

\begin{lstlisting}[language=bash]
# For your user's pip installation.
# Note! On MacOS, replace $HOME/.local/bin with $HOME/Library/Python/3.7/bin
export PATH=$PATH:$HOME/.local/bin
# For the Python modules in open_spiel.
export PYTHONPATH=$PYTHONPATH:/<path_to_open_spiel>
# For the Python bindings of Pyspiel
export PYTHONPATH=$PYTHONPATH:/<path_to_open_spiel>/build/python
\end{lstlisting}

\subsection{Running the First Example}
\label{sec:first-example}

After having built OpenSpiel following Sec~\ref{sec:building}, run the example from the \texttt{build} directory without any arguments:
\begin{lstlisting}[language=bash]
examples/example
\end{lstlisting}

\noindent This prints out a list of registered games and the usage. Now, let's play a game of Tic-Tac-Toe with uniform random players:
\begin{lstlisting}[language=bash]
examples/example --game=tic_tac_toe
\end{lstlisting}

Wow -- how exhilarating! Now, why not try one of your favorite games?

Note that the structure in the build directory mirrors that of the source, so the example is found in 
\texttt{open\_spiel/examples/example.cc}. At this stage you can run one of many binaries created, such as
\texttt{games/backgammon\_test} or \texttt{algorithms/external\_sampling\_mccfr\_test}.

Once you have set your PYTHONPATH as explained in Sec~\ref{sec:pythonpath}, you can similarly run
the python examples:
\begin{lstlisting}[language=bash]
cd ../open_spiel
python3 python/examples/example.py --game=breakthrough
python3 python/examples/matrix_game_example.py
\end{lstlisting}

Nice!

\subsection{Adding a New Game}

We describe here only the simplest and fastest way to add a new game. It is ideal to first be aware of the general API,
which is described on a high level in Section~\ref{sec:design-api}, on github, and via comments in \texttt{spiel.h}.

\begin{enumerate}
\item Choose a game to copy from in {\tt games/}. Suggested games: Tic-Tac-Toe and Breakthrough for perfect information without chance events,
Backgammon or Pig for perfect information games with chance events, 
Goofspiel and Oshi-Zumo for simultaneous move games, and Leduc poker and Liar's dice for imperfect information games.
For the rest of these steps, we assume Tic-Tac-Toe. 
\item Copy the header and source: 
\texttt{tic\_tac\_toe.h}, \texttt{tic\_tac\_toe.cc}, and \texttt{tic\_tac\_toe\_test.cc}
to \texttt{new\_game.h}, \texttt{new\_game.cc}, and \texttt{new\_game\_test.cc}.
\item Add the new game's source files to \texttt{games/CMakeLists.txt}.
\item Add the new game's test target to \texttt{games/CMakeLists.txt}
\item In \texttt{new\_game.h}, rename the header guard at the the top and bottom of the file.
\item In the new files, rename the inner-most namespace from \texttt{tic\_tac\_toe} to \texttt{new\_game}
\item In the new files, rename \texttt{TicTacToeGame} and \texttt{TicTacToeState} to \texttt{NewGameGame} and \texttt{NewGameState}
\item At the top of \texttt{new\_game.cc}, change the short name to \texttt{new\_game} and include the new game's header.
\item Add the short name to the list of expected games in \texttt{python/tests/pyspiel\_test.py}.
\item You should now have a duplicate game of Tic-Tac-Toe under a different name. It should build and the test should run, 
and can be verified by rebuilding and running the example from Section~\ref{sec:first-example}.
\item Now, change the implementations of the functions in \texttt{NewGameGame} and \texttt{NewGameState} to reflect your new game's 
logic. Most API functions should be clear from the game you copied from. If not, each API function that is overridden will be
fully documented in superclasses in \texttt{spiel.h}. See also the description of extensive-form games in Section~\ref{sec:efg}
which closely matches the API.
\item Once done, rebuild and rerun the tests from Sec~\ref{sec:building} to ensure everything
passes (including your new game's test!)
\end{enumerate}

\subsection{Adding a New Algorithm}

Adding a new algorithm is fairly straight-forward. Like adding a game, it is easiest to copy and start
from one of the existing algorithms. If adding a C++ algorithm, choose one from \texttt{algorithms/}. If adding
a Python algorithm, choose one from \texttt{python/algorithms/}. For appropriate matches, see Table~\ref{tab:algorithms}.

Unlike games, there is no specific structure or API that must be followed for an algorithm. If the algorithm is one in a class of
existing algorithms, then we advise keeping the style and design similar to the ones in the same class,  re-using
function or modules where possible.

The algorithms themselves are not binaries, but classes or functions that can be used externally.
The best way to show an example of an algorithm's use is via a test. However, there are also binary executables in 
\texttt{examples/} and \texttt{python/examples/}.

\section{Design and API}
\label{sec:design-api}

The purpose of OpenSpiel is to promote {\it general} multiagent reinforcement learning across many different game types,
in a similar way as general game-playing~\cite{GGP} but with a heavy emphasis on learning and not in competition form.
We hope that OpenSpiel could have a similar effect on general RL in games as the 
Atari Learning Environment~\cite{bellemare13arcade,machado18arcade}
has had on single-agent RL. 

OpenSpiel provides a general API with a C\texttt{++} foundation, which is exposed through Python bindings 
(via \texttt{pybind11}). Games are
written in C\texttt{++}. This allows for fast or memory-efficient implementations of basic algorithms that might
need the efficiency. Some custom RL environments are also implemented in Python. 
Most algorithms that require machine learning are implemented in Python.

Above all, OpenSpiel is designed to be easy to install and use, easy to understand, easy to extend (``hackable''),
and general/broad. OpenSpiel is built around two major important design criteria:

\setuldepth{bla}
\begin{enumerate}
\item {\color{red}{\bf \ul{Keep it simple}}}. Simple choices are preferred to more complex ones. The code should be readable, usable, extendable
by non-experts in the programming language(s), and especially to researchers from potentially different fields. OpenSpiel provides
reference implementations that are used to learn from and prototype with, rather than fully-optimized / high-performance code
that would require additional assumptions (narrowing the scope / breadth) or advanced (or lower-level) language features.
\item {\color{red}{\bf \ul{Keep it light}}}. Dependencies can be problematic for long-term compatibility, maintenance, and ease-of-use. Unless there
is strong justification, we tend to avoid introducing dependencies to keep things portable and easy to install.
\end{enumerate}

\subsection{Extensive-Form Games}
\label{sec:efg}

There are several formalisms and corresponding research communities for representing multiagent interactions.
It is beyond the scope of this paper to survey the various formalisms, so we describe the ones most relevant to our
implementations. There have been recent efforts to harmonize the terminology and make useful associations among
algorithms between computational game theory and reinforcement learning~\cite{Srinivasan18RPG,Lockhart19ED,Kovarik19FOG},
so we base our terminology on classical concepts and these recent papers.

Games in OpenSpiel are represented as procedural extensive-form games~\cite{OsbRub94,Shoham09}, though in some cases can also
be cyclic such as in Markov Decision Processes~\cite{Sutton18} and Markov games~\cite{Littman94markovgames}. We first give
the classical definitions, then describe some extensions, and explain some equivalent notions between the fields of
reinforcement learning and games.

An \defword{extensive-form game} is a tuple $\langle \cN, \cA, \cH, \cZ, u, \tau, \cS \rangle$, where
\begin{itemize}
    \item $\cN = \{1, 2, \ldots n\}$ is a finite set of $n$ \defword{players}\footnote{Note that the player IDs range from $0$ to $n-1$ in the implementations.}. There is also a special player $c$, called \defword{chance}.
    \item $\cA$ is a finite set of \defword{actions} that players can take. This is a global set of state-independent actions; generally, only a subset of {\it legal} actions are available when agents decide.
    \item $\cH$ is a finite set of \defword{histories}. Each history is a sequence of actions that were taken from the start of the game.
    \item $\cZ \subseteq \cH$ is a subset of \defword{terminal histories} that represents a completely played game.
    \item $u : \cZ \rightarrow \Delta_u^n \subseteq \Re^n$, where $\Delta_u = [u_{\min}, u_{\max}]$, is the utility function assigning each player a utility at terminal states, and $u_{\min}, u_{\max}$ are constants representing the minimum and maximum utility. 
    \item $\tau : \cH \rightarrow \cN$ is a \defword{player identity} function; $\tau(h)$ identifies which player acts at $h$.
    \item $\cS$ is a set of \defword{states}. In general, $\cS$ is a partition of $\cH$ such that each state $s \in \cS$ contains histories $h \in s$ that cannot be distinguished by $\tau(s) = \tau(h) \mbox{ where } h \in s$. Decisions are made by players at these states. There are several ways to precisely define $\cS$ as described below.
\end{itemize}

We denote the legal actions available at state $s$ as $\cA(s) \subseteq \cA$.
Importantly, a history represents the true ground/world state: when agents act, they change this history, but depending on how
the partition is chosen, some actions (including chance's) may be private and not revealed to some players.

We will extend this formalism further on to more easily describe how games are represented in OpenSpiel. However, we can
already state some important categories of games:
\begin{itemize}
    \item A \defword{constant-sum} ($k$-sum) game is one where $\forall z \in \cZ, \sum_{i \in \cN} u_i(z) = k$. 
    \item A \defword{zero-sum} game is a constant-sum game with $k = 0$.
    \item An \defword{identical interest} game is one where $\forall z \in \cZ, \forall i,j \in \cN, u_i(z) = u_j(z)$.
    \item A \defword{general-sum game} is one without any constraint on the sum of the utilities.
\end{itemize}
In other words: $k$-sum games are strictly competitive, identical interest games are strictly cooperative, and general-sum games
are neither or somewhere in between. Also,
\begin{itemize}
    \item A \defword{perfect information} game is one where there is only one history per state: $\forall s \in \cS, |s| = 1$. 
    \item A \defword{imperfect information} game is one where there is generally more than one history per state, $\exists s \in \cS: |s| > 1$.
\end{itemize}
Chess, Go, and Breakthrough are examples of perfect information games without events (no chance player).
Backgammon and Pig are examples of perfect information games with chance events.
Leduc poker, Kuhn poker, Liar's Dice, and Phantom Tic-Tac-Toe are examples of imperfect information games.
Every one of these example games is zero-sum.

\begin{definition}
A \defword{chance node} (or chance event) is a history $h$ such that $\tau(h) = c$.
\end{definition}
In zero-sum perfect information games, minimax and alpha-beta search are classical search algorithms for making decisions
using heuristic value functions~\cite{KnuthMoore75}.
The analogs for perfect information games with chance events are expectiminimax~\cite{Michie66} and *-minimax~\cite{Ballard83}.

\subsubsection{Extension: Simultaneous-Move Games}

We can augment the extensive-form game with a special kind of player, the simultaneous move player: $\div$.
When $\tau(s) = \div$, each player $i$ has a set of legal actions $\cA_i(s)$, and all players act simultaneously
choosing a \defword{joint action} $a = (a_i)_{\{i \in \cN\}}$.
Histories in these games are then sequences of joint actions, and transitions take the form $(h, a, h')$.
The rest of the properties from extensive-form games still hold.
\begin{definition}
A \defword{normal-form} (or \defword{one-shot} game) is a simultaneous-move game with a single state, $|S| = 1$.
A \defword{matrix game} is a normal-form game where $|\cN| = 2$.
\end{definition}
\begin{fact}
A simultaneous-move game can be represented as a specific type of extensive-form game with imperfect information.
\end{fact}
To see why this is true: consider the game of Rock, Paper, Scissors ($\cA = \{ \textsc{r}, \textsc{p}, \textsc{s} \}$)
where each player chooses a single action, revealing their choice simultaneously. An equivalent turn-based is the
following: the first player
writes their action on a piece of paper, and places it face down. Then, the second player does the same. Then,
the choices are revealed simultaneously. The players acted at separate times, but the second player did not know
the choice made by the first player (and hence could be in one of three histories: 
$h = \textsc{r}, h = \textsc{p}, \mbox{ or } h = \textsc{s}$), and the game has two states instead of one state.
In a game with many states, the same idea can simply be repeated for every state.

Why, then, represent these games differently? There are several reasons:
\begin{enumerate}
    \item They have historically been treated as separate in the multiagent RL literature. 
    \item They can sometimes be solved using Bellman-style dynamic programming, unlike general imperfect information games.
    \item They are slightly more general. In fact, one can represent a turn-based game using a simultaneous-move game, simply by setting
          $\cA_i(s) = \emptyset$ for $j \not= \tau(s)$ or by adding a special $\textsc{pass}$ move as the only legal action
          when it is not a player's turn.
\end{enumerate}
We elaborate on each of these points in the following section, when we relate simultaneous-move games to existing
multiagent RL formalisms.

\subsubsection{Policies, Objectives, and Multiagent Reinforcement Learning}

We now add the last necessary ingredients for designing decision-making and learning algorithms, and bring in 
the remaining standard RL terms.
\begin{definition}
A \defword{policy} $\pi : \cS \rightarrow \Delta(\cA(s))$, where $\Delta(X)$ represents the set of probability
distributions over $X$, describes agent behavior. An agent acts by selecting actions from its policy: $a \sim \pi$.
A \defword{deterministic} policy is one where at each state the distribution over actions has probability $1$ on
one action and zero on the others. A policy that is not (necessarily) deterministic is called \defword{stochastic}.
\end{definition}
In games, the chance player is special because it {\it always plays with a fixed (stochastic) policy $\pi_c$}.

\begin{definition}
A \defword{transition function} $\cT : \cS \times \cA \rightarrow \Delta(\cS)$ defines a probability distribution
over successor states $s'$ when choosing action $a$ from state $s$.
\end{definition}
\begin{fact}
A transition function can be equivalently represented using intermediate chance nodes between the histories of the
predecessor and successor states $h \in s$ and $h' \in s'$. The transition function is then determined by $\pi_c$ and
$\Pr(h | s)$.
\end{fact}
\begin{definition}
A player, or agent, has \defword{perfect recall} if, the state does not lose the information about the past
decisions made by the player. Formally, all histories $h \in s$, contain the same sequence of action of the 
current player: let $\textsc{SAHist}_i(h)$ be the history of only player $i$'s state-action pairs $(s,a)$
experienced along $h$.
Player $i$ has perfect recall if for all
$s \in \{ s~|~s \in \cS, \tau(s) = i \}, \mbox{ and all } h, h' \in s, \textsc{SAHist}_i(h) = \textsc{SAHist}_i(h')$.
\label{def:perfect-recall}
\end{definition}
In Poker, a player acts from an information state, and the histories corresponding to such an information state
only differ
in the chance event outcomes that correspond to the opponent's
private cards. In these partially-observable games, a state is normally called an \defword{information state} to
emphasize the fact that the agent's perception of the state ($s$) is different than the true underlying world state
(one of $h \in s$).

The property of perfect recall turns out to be a very important criterion for determining convergence guarantees for
exact tabular algorithms, as we show in Section~\ref{sec:alg-results}.

\begin{definition}
An observation is a partial view of the information state and contains strictly less information than
the information state. To be valid, the sequence of observations and actions of all players should contain
at least as much information as the information state. Formally: Let $\Omega$ be a finite set of
\defword{observations}. Let $O_i: \cS \rightarrow \Omega$ be an observation
function for player $i$ and denote $o_i(s)$ as the observation. As $s$ contains histories $h$, we will
write $o_i(h) = o_i(s)$ if $h \in s$. A valid observation is such that the function
$h \rightarrow (o_i(h'))_{h'\sqsubset h}$ defines a partition of the history space $\cH$ that is a
sub-partition of $\cS$.
\end{definition}

In a multiplayer game, we define a per-step \defword{reward} {\it to player $i$} for a transition as $r_i(s,a,s')$,
with $r(s, a, s')$ representing the vector of returns to all players.
In most OpenSpiel games, these $r(s, a, s') = 0$ until $s'$ is terminal, ending the episode, and these values
are obtained by {\tt State::Rewards} and {\tt State::PlayerReward} function called on $s'$.
Player interaction over an episode generates a trajectory $\rho = (s_0, a_0, s_1, \cdots)$ whose length
is $|\rho|$. 
We define a \defword{return} {\it to player $i$} as $g^{\rho}_{t,i} = \sum_{t' \ge t}^{|\rho|-1} r_i(s_{t'}, a_{t'}, s_{t'+1})$
with $g^\rho_t$ representing a vector of rewards to all players as with per-step rewards.
In OpenSpiel, the {\tt State::Returns} function provides $g^{\rho}_0$ and {\tt State::PlayerReturn} provides $g^\rho_{0,i}$.
Note that we do not use a discount factor when defining rewards here because most games are episodic; learning agents are 
free to discount rewards however they like, if necessary.
Note also that the standard (undiscounted) return is the random variable $G_t$.

Each agent's \defword{objective} is to maximize its own return, $G_{0,i}$ or an {\it expected return}
$\bE_{z \sim \pi}[G_0,i]$. However, note that the trajectory sampled depends not just on player $i$'s 
policy but on {\it every other player's policies}! So, an agent cannot maximize its return in isolation:
it {\it must} consider the other agents as part of its optimization problem. This is fundamentally different
from traditional (single-agent) reinforcement learning, and the main challenge of multiagent RL.

\subsection{Algorithms and Results}
\label{sec:alg-results}

\subsubsection{Basic Algorithms}

Suppose players are playing with a joint policy $\pi$.
The expected returns algorithm computes $\bE_{\pi}[G_{0,i}]$ for all players $i \in \cN$
exactly, by doing a tree traversal over the game and querying the policy at each state $s$.
Similarly, for small enough games, one can get all the states ($\cS$) in a game by doing
a tree traversal and indexing each state by its information state string description.

The trajectories algorithms run a batch of episodes by following a joint policy $\pi$,
collecting various data such as the states visited, state policies, actions sampled, returns,
episode lengths, etc., which could form the basis of the data collection for various RL algorithms.

There is a simple implementation of value iteration. In single-agent games, it is identical
to the standard algorithm~\cite{Sutton18}. In two-player turn-taking zero-sum games, the
values for state $s$, i.e. $V(s)$, is stored in view of the player to play at $s$,
i.e. $V_{\tau(s)}(s)$. This can be solved by applying the identities $V_1(s) = -V_2(s)$ and
$r_1(s,a,s') = -r_2(s,a,s')$.

\subsubsection{Search Algorithms}

There are two classical search algorithms for zero-sum turn-taking games of perfect information:
minimax (and alpha-beta) search~\cite{KnuthMoore75,AIbook}, and Monte Carlo tree search (MCTS)~\cite{Coulom06,UCT,mctssurvey}. 

Suppose one wants to choose at some root state $s_{root}$ : given a heuristic value function for 
$v_{0,i}(s)$ (representing the value of state $s$ to player $i$) and some depth $d$, minimax 
search computes a policy $\pi(s)$ that assigns 1 to an action that
maximizes the following depth-limited adversarial multistep value backup:
\[
v_{d}(s) = \left\{ \begin{array}{ll}
                   v_{0,\tau{s_{root}}}(s)  & \mbox{if $d = 0$}; \\
                   \max_{a \in \cA(s)} v_{d-1}(\cT(s, a)) & \mbox{if $\tau(s) = i$}; \\
                   \min_{a \in \cA(s)} v_{d-1}(\cT(s, a)) & \mbox{if $\tau(s) \not= i$}, \end{array} \right.
\]
where here we treat $\cT(s,a) = s'$ as a deterministic map for the successor state reached from
taking action $a$ in state $s$.

The Python implementation of minimax includes expectiminimax~\cite{Michie66} as well, which also backs up
expected values at chance nodes. Alpha-beta style cut-offs could also be applied using $*$-minimax~\cite{Ballard83},
but it is not currently implemented.

The implementations of MCTS are vanilla UCT with random playouts. Chance node are supported and represented
explicitly in the tree: at chance nodes, the tree policy is always to sample according to the chance node's
probability distibution.

\subsubsection{Optimization Algorithms}

OpenSpiel includes some basic optimization algorithms applied to games, such as solving zero-sum matrix
games (\cite[Section 4]{Shoham09}, \cite{Littman94markovgames}) and sequence-form linear
programming for two-player zero-sum extensive-form games 
(\cite{SequenceFormLPs} and \cite[Section 5]{Shoham09}), and an algorithm to check whether an action
is dominated by a mixture of other strategies in a normal-form~\cite[Sec 4.5.2]{Shoham09}.

\subsubsection{Traditional Single-Agent RL Algorithms}

We currently have three algorithms usable for traditional (single-agent) RL:
Deep Q-Networks (DQN)~\cite{Mnih15DQN},
Advantage Actor-Critic (A2C)~\cite{Mnih2016asynchronous},
and Ephemeral Value Adjustments (EVA)~\cite{Hansen18EVA}.
Each algorithm will operate as the standard one in single-agent environments.

Each of these algorithms can also be run in the multiagent setting, in various ways.
The default is that each player is independently running a copy of the algorithm with states and
observations that include what other players did. The other way to use these algorithms is to
compute an approximate best response to a fixed set of other players' policies, described in
Section~\ref{sec:po-games}.

The main difference between the implementations of these algorithms and other standard ones is that
these are aware that only a subset of actions are legal / illegal. So, for example, in Q-learning
the value update for a transition $(s, a, s')$ and policy updates are:
\begin{equation}
Q(s,a) \leftarrow Q(s,a) + \alpha(r + \gamma \max_{a' \in \cA(s')} Q(s', a') - Q(s,a)),
\end{equation}
\begin{equation}
\pi(s,a) = \left\{ \begin{array}{ll}
                   0                                         & \mbox{ if } a \not\in \cA(s); \\
                   1 - \epsilon + \frac{\epsilon}{|\cA(s)|}  & \mbox{ if } a = \argmax_{a' \in \cA(s)}Q(s,a'); \\
                   \frac{\epsilon}{|\cA(s)|}                 & \mbox{otherwise}. \end{array} \right.\\
\end{equation}
Note that the actions are in the set of {\it legal actions} $\cA(s)$ and $\cA(s')$ rather than assuming
that every
action is legal at every state. For policy gradient methods, a masked softmax is used to set the
logits of the illegal actions to $-\infty$ to force the policy to sets probability zero to
illegal actions.

\subsubsection{Partially-Observable (Imperfect Information) Games}
\label{sec:po-games}

There are many algorithms for reinforcement learning in partially-observable (zero-sum) games,
as this is the focus of the core team's research interests.

{\bf Best Response and NashConv}

Suppose $\pi$ is a joint policy. A \defword{best response} policy for player $i$ is a policy that
maximized player $i$'s return against the other players' policies ($\pi_{-i}$). There may be many
best responses, and we denote the set of such best responses, 
\[ BR(\pi_{-i}) = \{ \pi_i'~|~ \pi_i' = \argmax_{\pi_i} u_i(\pi_i, \pi_{-i}) \}. \]
Let $\delta_i(\pi)$ be the incentive for player $i$ to deviate to one of its best responses: 
$\delta_i(\pi) = u_i(\pi_i^b, \pi_{-i}) - u_i(\pi), $ where $\pi_i^b \in BR(\pi_{-i})$.
An approximate \defword{${\bm \epsilon}$-Nash equilibrium} is a joint policy such that
$\delta_i(\pi) \le \epsilon$ for all $i \in \cN$, where a Nash equilibrium is obtained at $\epsilon = 0$.

A common metric for determining the rates of convergence (to equilibria) of algorithms in practice is:
\[ \textsc{NashConv}(\pi) = \sum_{i \in \cN} \delta_i(\pi). \]
In two-player constant-sum (i.e. $k$-sum) games, a similar metric has been used:
\[
\textsc{Exploitability}(\pi) = \frac{\textsc{NashConv}(\pi)}{|\cN|}
                             = \frac{\sum_{i \in \cN} \delta_i(\pi)}{n}
                             = \frac{u_1(\pi_1^b, \pi_2) + u_2(\pi_1, \pi_2^b) - k}{2},
\]
where $\pi_i^b \in BR(\pi_{-i})$. Nash equilibria are often considered optimal in two-player
zero-sum games, because they guarantee
maximal worst-case returns against any other opponent policy. This is also true for approximate
equilibria, so convergence to equilibra has been a focus in this class of games.

{\bf Fictitious Play and Best Response-Based Iterative Algorithms}

Fictitious play (FP) is a classic iterative procedure for computing policies
in (normal-form) games~\cite{Brown51,Robinson51}.
Starting with a uniform random policy at time $t = 0$. Then, for $t \in \{ 1, 2, \cdots \}$, do:
\begin{enumerate}
\item Each player computes a best response to the opponents' average policy:
      $\pi^t_i \in BR(\bar{\pi}^{t-1}_{-i})$.
\item Each player updates their average policy: $\bar{\pi}^t_i = \frac{(t-1) \bar{\pi}^{t-1}_i + \pi^t_i}{t}$.
\end{enumerate}

OpenSpiel has an implementation of extensive-form fictitious play (XFP)~\cite{Heinrich15FSP}, which is
equivalent to the classical fictitious play. To run it on normal-form games, the game needs to be transformed
into a turn-based game using {\tt TurnBasedSimultaneousGame} in {\tt game\_transforms/}.
Fictitious Self-Play is a sampled-based RL version of XFP that uses supervised learning to learn the average
policy and reinforcement learning to compute approximate best responses. Neural Fictitious
Self-Play (NFSP) scales these ideas using neural networks and a reservoir-sampled buffer to maintain a
uniform sample of experience to train the average policy~\cite{Heinrich16}.

The average policy in fictitious play can be described equivalently as a meta-policy that assigns uniform
weight over all the previous best response
policies, and each iteration computes a best response to the opponents' meta-policies.
Policy-Space Response Oracles (PSRO) generalizes fictitious play and the double-oracle
algorithm~\cite{Lanctot17PSRO,McMahan03Planning} by analyzing this meta-game using 
empirical game-theoretic analysis~\cite{Wellman06}.
Exploitabiliy Descent replaces the second step of fictitious play with a policy gradient ascent against the
state-action values given the opponents play their best responses~\cite{Lockhart19ED}. This one change allows
convergence of the policies themselves rather than having to maintain an average policy; in addition, it 
makes the optimization of the polices amenable to RL-style general function approximation.

\begin{figure}[h!]
    \centering
    \includegraphics[scale=0.6]{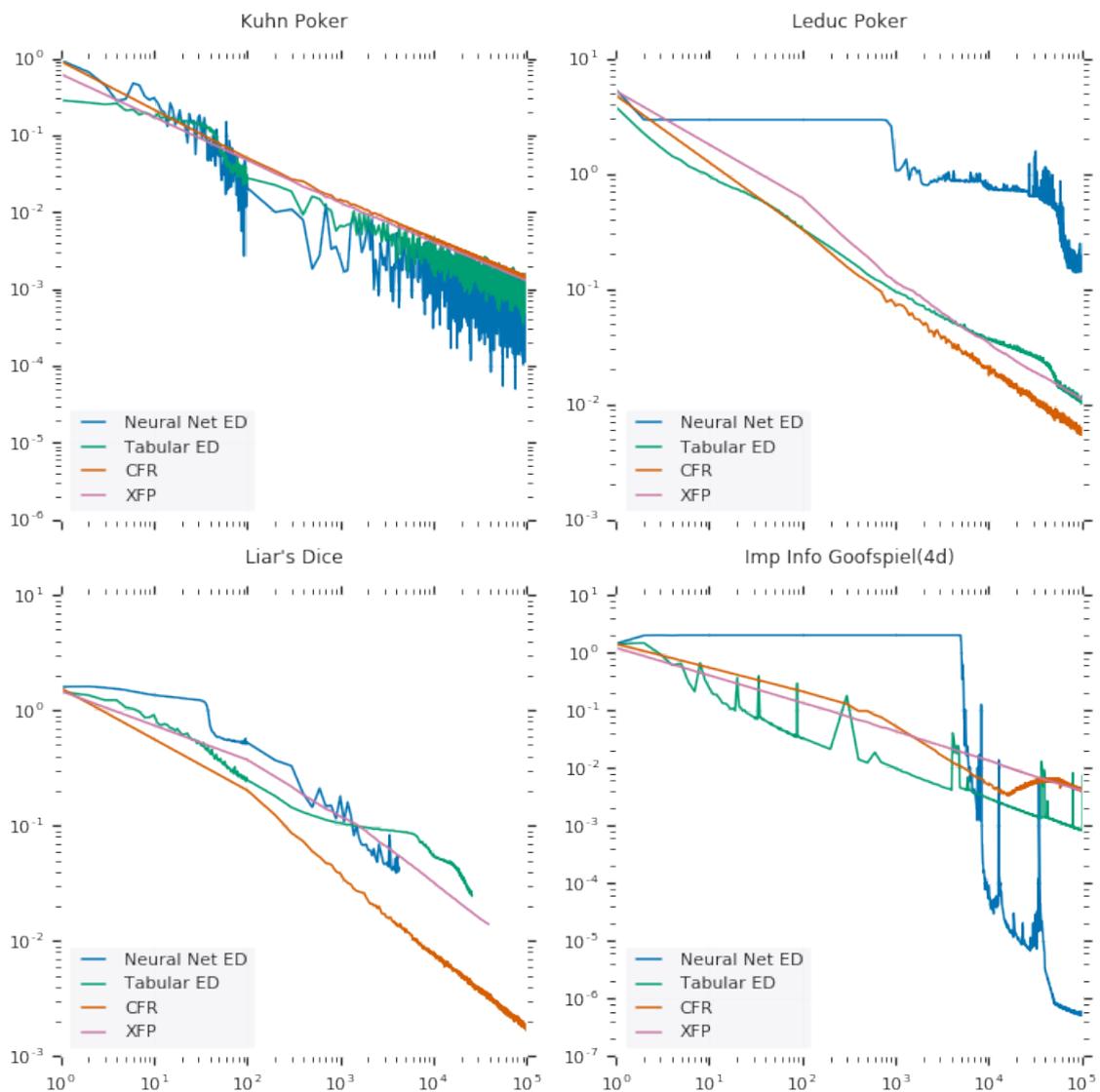}
    \caption{Convergence rates of XFP and ED algorithms on various partially-observable games in OpenSpiel.
    The units of the $x$-axis is iterations and the units of the $y$-axis is $\textsc{NashConv}$. Figure
    taken from~\cite{Lockhart19ED}.}
    \label{fig:psro-ed-convergence-graphs}
\end{figure}

\begin{figure}[h!]
    \centering
    \includegraphics[scale=0.4]{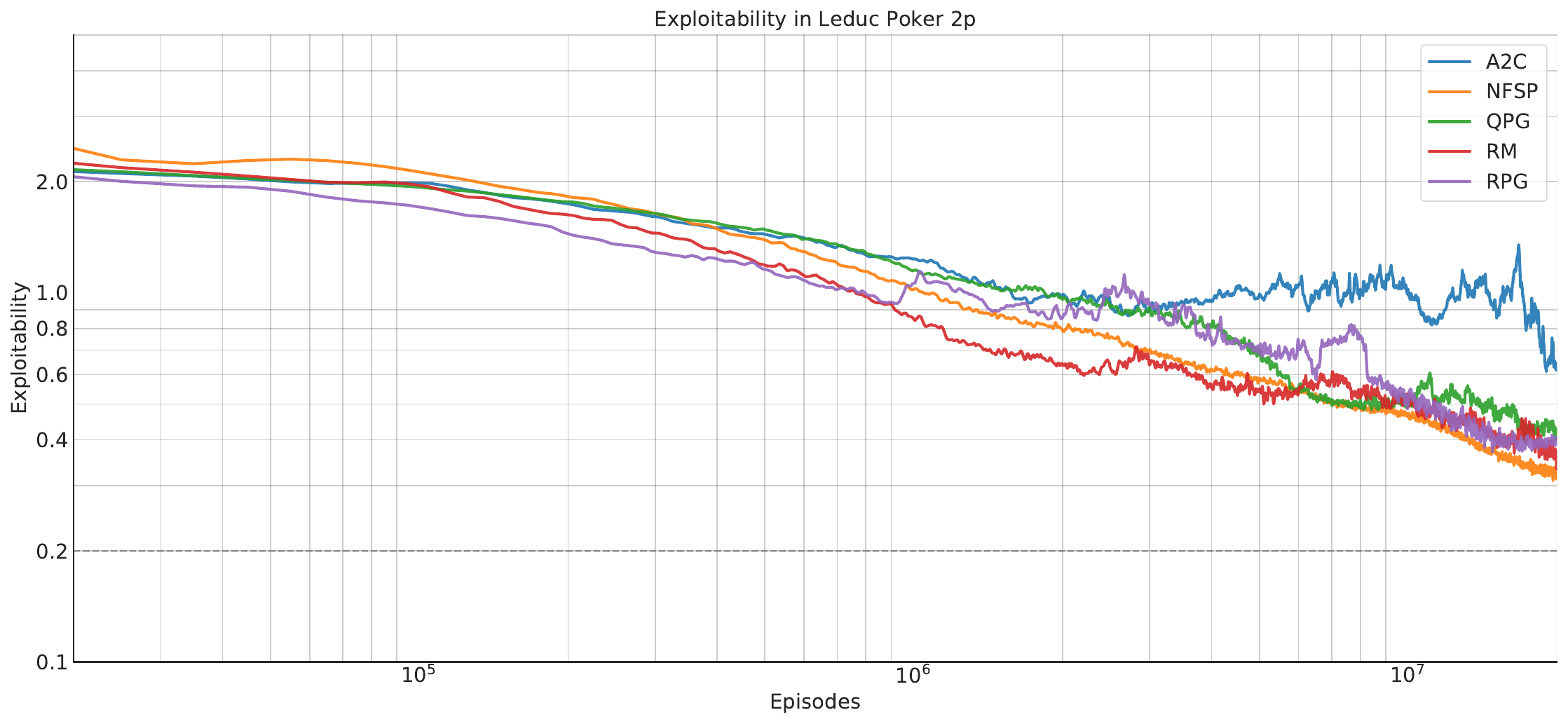}
    \caption{Convergence rates of NFSP and various (regret-based) policy gradient algorithms in 2-player Leduc poker.
    Each line is an average over the top five seeds and hyperparemeter settings for each algorithm. The lowest (around 0.2) exploitability value reached by any individual run is depicted by a dashed line.}
    \label{fig:nfsp-rpg-leduc}
\end{figure}

A convergence curve for XFP and ED are shown in Figure~\ref{fig:psro-ed-convergence-graphs}. 
A convergence curve for NFSP in 2-player Leduc is found below (Figure~\ref{fig:nfsp-rpg-leduc}), included with the policy gradient methods.

\vspace{2cm}


\vspace{1cm}

{\bf Counterfactual Regret Minimization}

Counterfactual regret (CFR) minimization is a policy iteration algorithm for computing approximate
equilibra in two-player
zero-sum games~\cite{CFR}. It has revolutionized Poker AI research~\cite{Rubin11Poker,Sandholm10The},
leading to the largest variants of poker being solved and competitive polices that have beat top human
professionals~\cite{Bowling15Poker,Moravcik17DeepStack,Brown17Libratus,Brown19}.

CFR does two main things: (a) define a new notion of state-action value, the counterfactual
value, and
(b) define a decomposed regret minimization procedure (based on these values) at every information state
that, together, 
leads to minimization of overall average regret. This means that the average policy of two CFR players
approaches an approximate equilibrium.

Define $\cZ(s)$ as the set of terminal histories that pass through $s$, paired with the prefix of each
terminal $h \sqsubset z$. Define a reach probability $\eta^{\pi}(h)$ to be the product of all players'
probabilities of state-action pairs along $h$ (including chance's), which can be decomposed into 
player $i$'s contribution and their opponents' contributions: $\eta^{\pi}(h) = \eta_i^\pi(h) \eta_{-i}^\pi(h)$.
Similarly define $\eta^\pi(h,z)$ similarly from $h$ to $z$ and $ha$ as the history $h$ appended with action $a$.
The counterfactual state-action value for $i = \tau(s)$ is:
\[ q^c_{\pi,i}(s,a) = \sum_{(h,z) \in \cZ(s)} \eta_{-i}^{\pi}(h) \eta^{\pi}(ha,z) u_i(z). \]
The state value is then $v_{\pi,i}^c(s) = \sum_{h \in s} \pi(s,a) q_{\pi,i}^c(s,a)$.

CFR starts with a uniform random policy $\pi^0$ and proceeds by applying regret minimization at every
information state independently. Define $r^t(s,a) = q^c_{\pi^t,i}(s,a) - v^c_{\pi^t,i}(s)$ to be the instantaneous
\defword{counterfactual regret}. CFR proceeds by minimizing this regret, typically using
regret-matching~\cite{Hart00}. A table of cumulative regret is maintained $R^t(s,a) = \sum_t r^t(s,a)$, and the
policy at each state is updated using:
\[ \pi^{t+1}(s,a) =  \left\{ \begin{array}{ll}
                     \frac{R^{t,+}(s,a)}{\sum_{a \in \cA(s)}R^{t,+}(s,a)} & \mbox{if the denominator is positive}; \\
                        & \\
                     \frac{1}{|\cA(s)|} & \mbox{otherwise}, \end{array} \right.\\
\]
where $x^+ = \max(x,0)$.

In addition to basic CFR, OpenSpiel contains a few variants of Monte Carlo CFR~\cite{Lanctot09Sampling} such as outcome sampling
and external sampling, and CFR+~\cite{Tammelin15CFRPlus}.

{\bf Regression CFR}

Regression CFR (RCFR) was the first variant to combine RL-style function approximation with CFR
techniques~\cite{Waugh15solving,Morrill16}. The main idea is to train a regressor to predict the cumulative
or average counterfactual regrets, $\hat{R}^t(s,a) \approx R^t(s, a)$ or $\bar{r}'^t(s,a) \approx \nicefrac{R^t(s, a)}{t}$, instead of reading them from a table. The original paper
used domain-specific features and regression trees. The implementation in OpenSpiel uses neural networks with
raw inputs obtained by each game's {\tt InformationSetAsNormalizedVector} bit string.

\begin{figure}[h!]
    \centering
    \includegraphics[scale=0.8]{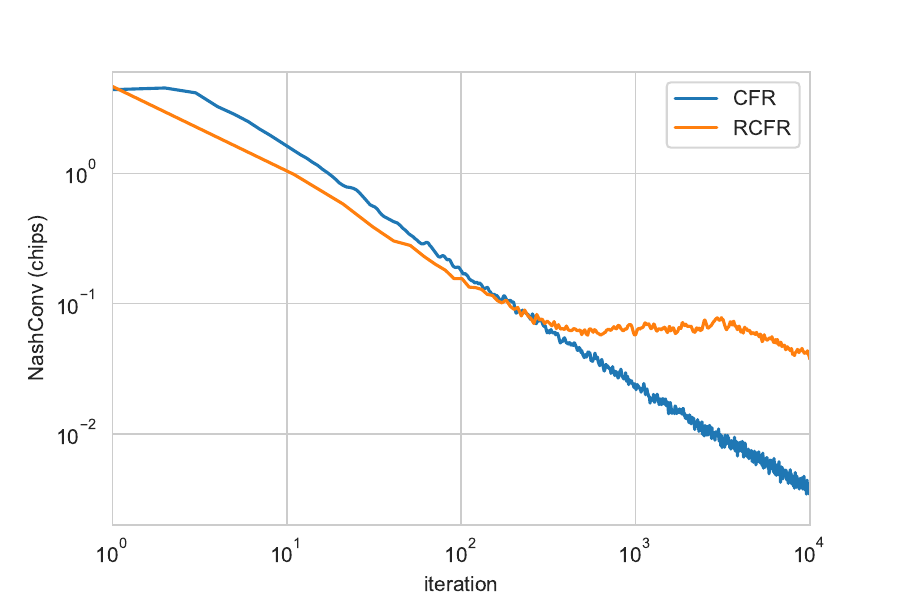}
    \caption{Convergence rate of RCFR in Leduc poker using a 2-layer network with 400 hidden units in each layer.
    The average policy is computed exactly (i.e. tabular), and regression targets are the cumulative predicted
    regrets.}
    \label{fig:rcfr}
\end{figure}

Figure~\ref{fig:rcfr} shows the convergence rate of RCFR compared to a tabular CFR.

Deep CFR~\cite{Brown19DeepCFR} applies these ideas to a significantly larger game using convolutional networks,
external sampling Monte Carlo CFR, and--like NFSP--a reservoir-sampled buffer.

{\bf Regret Policy Gradients}

Value-based RL algorithms, such as temporal-difference learning and Q-learning, {\it evaluate} a policy $\pi$ by
computing or estimating state (or state-action) values that represent the expected return conditioned on
having reached state $s$,
\[ v_{\pi}(s_t) = \bE_{\pi}[ G_t | S_t = s]. \]
Policies are {\it improved} by choosing the actions that lead to higher-valued states or higher-valued
returns.

In episodic partially-observable games, when agents have perfect recall (Def~\ref{def:perfect-recall}),
there is an important connection between
traditional values in value-based RL and counterfactual values~\cite[Section 3.2]{Srinivasan18RPG}:
\[ v_{\pi,i}(s) = \frac{v^c_{\pi,i}(s)}{\beta_{-i}(\pi, s)}, \]
where $\beta_{-i}(s) = \sum_{h \in s} \eta_{-i}^\pi(h)$ is the Bayes normalization term to ensure that 
$\Pr(h|s)$ is a probability distribution. CFR is then as a (tabular) all-actions
policy gradient algorithm with generalized infinitesimal gradient ascent (GIGA) at each
state~\cite{Srinivasan18RPG}, inspiring new RL variants for partially observable games.

These variants: Q-based ``all-actions'' Policy Gradient (QPG), Regret Policy Gradients (RPG),
and Regret-Matching Policy Gradients (RMGP) are included in OpenSpiel, along with classic batched A2C.
RPG differs from QPG in that the policy is optimized toward a no-regret region, minimizing the
loss based on $r^{+}(s,a)$, the motivation being that a policy with zero regret is, by definition, an equilibrium
policy. Convergence results for these algorithms are shown in Figure~\ref{fig:nfsp-rpg-leduc}.

{\bf Neural Replicator Dynamics}

Neural Replicator Dynamics (NeuRD)~\cite{Omidshafiei19NeuRD} takes the policy gradient connection to
CFR a step further:  in~\cite{Srinivasan18RPG}, the relationship between policy gradients and CFR was possible via
GIGA~\cite{Zinkevich03Online}; however, this requires $\ell_2$ projections of policies after the gradient
step. NeuRD, on the other hand, works directly with the common softmax-based policy representations.
Instead of differentiating through the softmax as policy gradient does, NeuRD differentiates only with respect
to the logits. This is equivalent to updating the policy of a 
parameterized replicator dynamics from evolutionary game theory~\cite{HofbauerSigmund98,Sandholm10Population}
using an Euler discretization.
The resulting update reduces to the well-known multiplicative weights update
algorithm or Hedge~\cite{Freund95}, which minimizes regret.
Hence, NeuRD in partially-observable games can replace regret-matching in CFR and retain convergence guarantees
in the tabular case since that algorithm reduces to CFR with Hedge.

\begin{figure}[h!]
    \centering
    \includegraphics[scale=1.0]{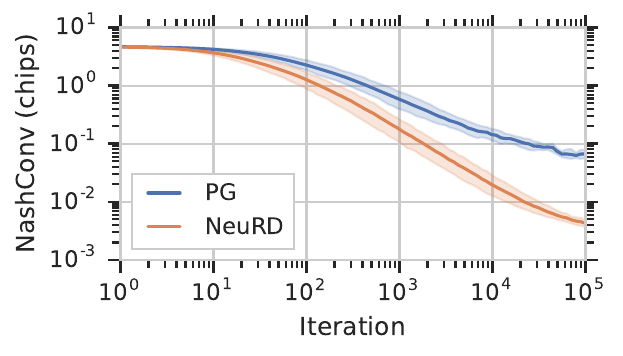}
    \caption{NashConv of tabular all-actions NeuRD versus tabular all-action policy gradient (policy gradient policy iteration)
    in Leduc poker. Figure taken from~\cite{Omidshafiei19NeuRD}.}
    \label{fig:neurd-tabular}
\end{figure}

\begin{figure}[h!]
    \centering
    \begin{tabular}{ccc}
    \includegraphics[width=0.3\textwidth]{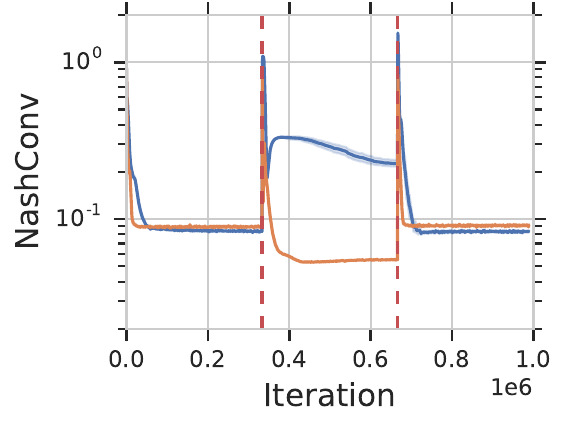} &
    \includegraphics[width=0.3\textwidth]{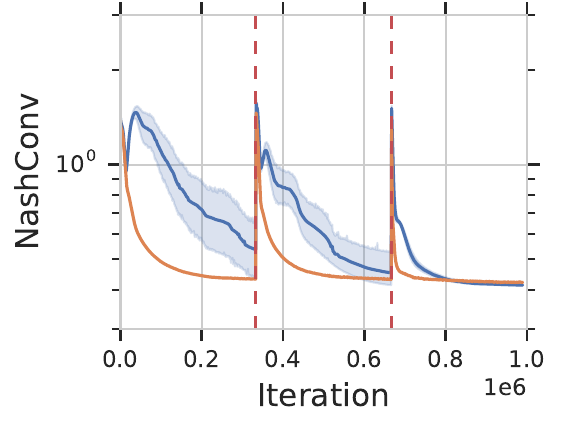} &
    \includegraphics[width=0.3\textwidth]{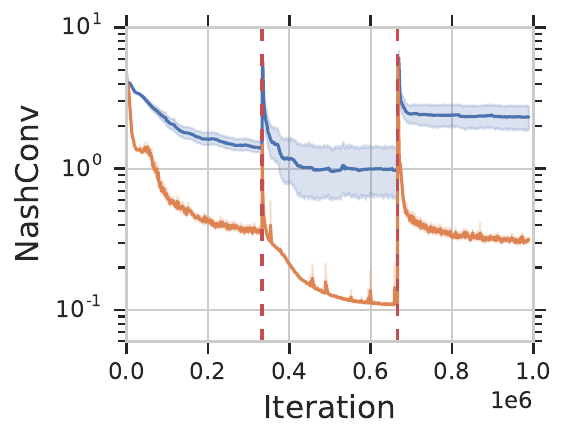} \\
    Kuhn poker & Goofspiel & Leduc poker \\
    \end{tabular}
    \caption{NashConv of NeuRD using sampling trajectories and function approximation. The games are played in three
    phases where, between phases, the returns are inverted. NeuRD is the yellow (bottom) line, which policy gradient
    is the blue (top) line. Figure taken from~\cite{Omidshafiei19NeuRD}.}
    \label{fig:neurd-nonstationary}
\end{figure}

One practical benefit is that the NeuRD policy updates are not weighted by the policy like policy gradient is. 
As a result, in non-stationary domains, NeuRD is also more adaptive to changes in the environment. Results for
NeuRD are show in Figures~\ref{fig:neurd-tabular} and \ref{fig:neurd-nonstationary}. 

\subsection{Tools and Evaluation}

OpenSpiel has a few tools for visualization and evaluation, though some would also be considered algorithms (such as $\alpha$-Rank). The best response algorithm is also a tool in some sense, but is listed in Section~\ref{tab:algorithms} due to its association with partially-observable games.

For now, all the tools and evaluation we mention in this section is contained under the {\tt python/egt} and {\tt python/visualizations} subdirectories of the code base.

\subsubsection{Visualizing a Game Tree}

\begin{figure}[h!]
    \centering
    \includegraphics[scale=0.7]{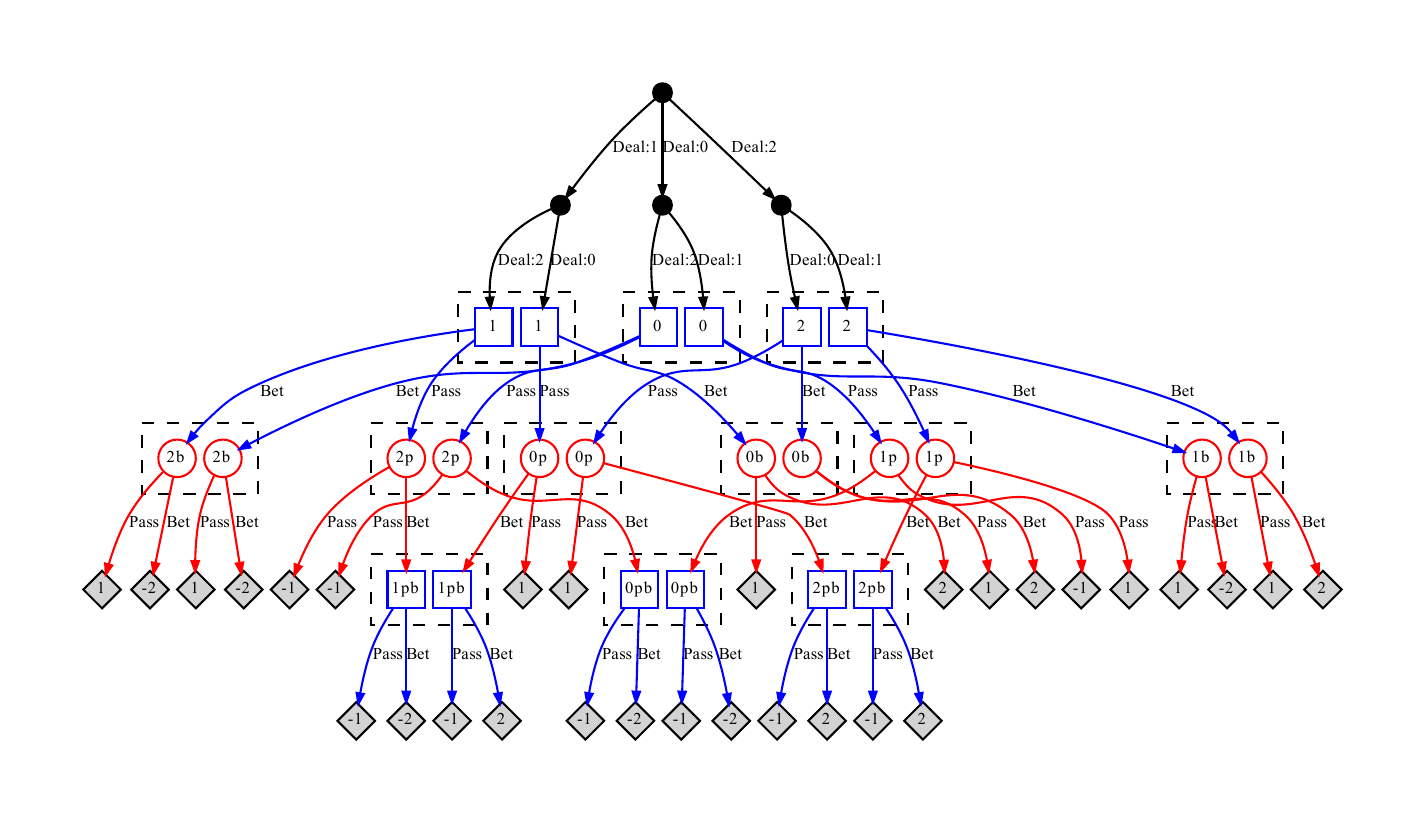}
    \caption{A visualization of Kuhn poker generated by {\tt python/examples/visualization\_example.py}.
    Black, blue, and red edges correspond to chance, first player, and second player outcomes/actions, respectively.
    Nodes correspond to histories $h$ and are labeled by their information state strings, and dotted boxes group these
    histories $h \in s$ by their information state $s$. Diamonds correspond to terminal states which are labeled by the
    utility to the first player.}
    \label{fig:kuhn-viz}
\end{figure}

A game tree can be visualized by using \href{http://www.graphviz.org}{Graphviz}. An example is shown in Fig~\ref{fig:kuhn-viz}.

\subsubsection{Visualization of Evolutionary and Policy Learning Dynamics}

One common visualization tool in the multiagent learning literature (especially in games) is a
\defword{phase portrait} that shows a vector field and/or trajectories of particle that depict local changes to the policy under specific update dynamics~\cite{Singh00,walsh:02,Bowling02,Walsh03,Bowling04,Wellman06,Abdallah08,Zhang10,Wunder2010,BloembergenTHK15,Tuyls18}.

For example, consider the well-known single-population replicator dynamic for symmetric games, where each player follows a learning dynamic
described by:
\[
\frac{\partial \pi_t(a)}{\partial t} = \pi_t(a) \left( u(a, \bm{\pi}_t) - \bar{u}(\bm{\pi}_t) \right)~~~~~\forall a \in \cA,
\]
where $u(a, \bm{\pi}_t)$ represents the expected utility of playing action $a$ against the full policy $\bm{\pi}_t$, 
and $\bar{u}(\bm{\pi}_t)$ is the expected value over all actions $\sum_{a \in \cA} \pi_t(a) u(a, \bm{\pi}_t)$.

\begin{figure}[h!]
    \centering
    \begin{tabular}{ccc}
      \includegraphics[scale=0.45]{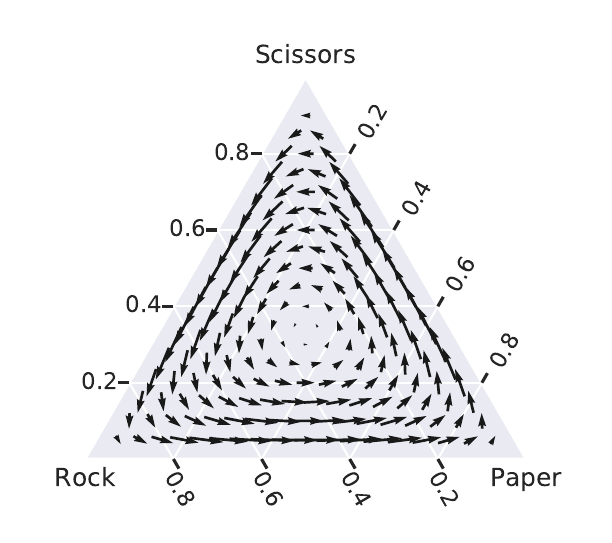} 
      \includegraphics[scale=0.45]{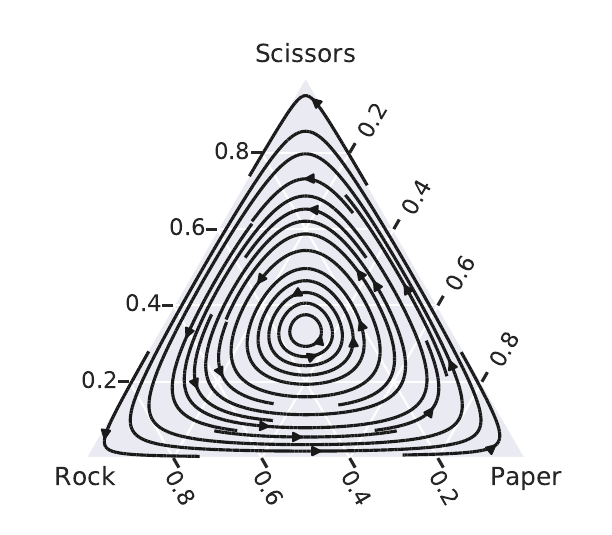} &
      \includegraphics[scale=0.45]{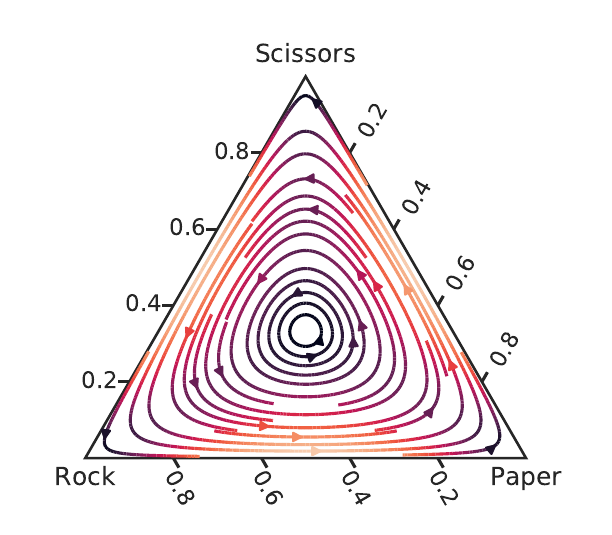} & 
    \end{tabular}
    \caption{Phase portraits of single-population replicator dynamics in \textit{Rock--Paper--Scissors}. The colored plot shows the relative magnitude of the dynamics.}
    \label{fig:rd-dynamics-rps}
\end{figure}

\begin{figure}[h!]
    \centering
    \begin{tabular}{cccc}
      \includegraphics[scale=0.45]{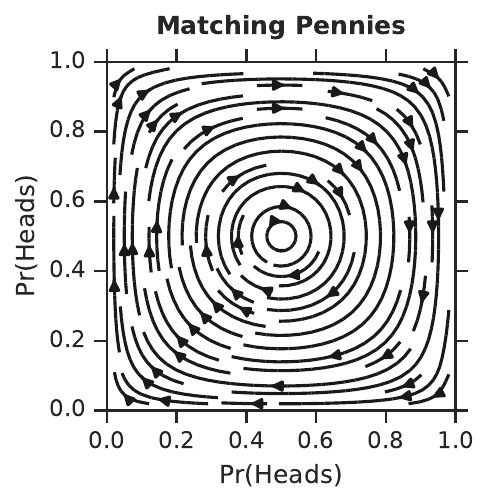} &
      \includegraphics[scale=0.45]{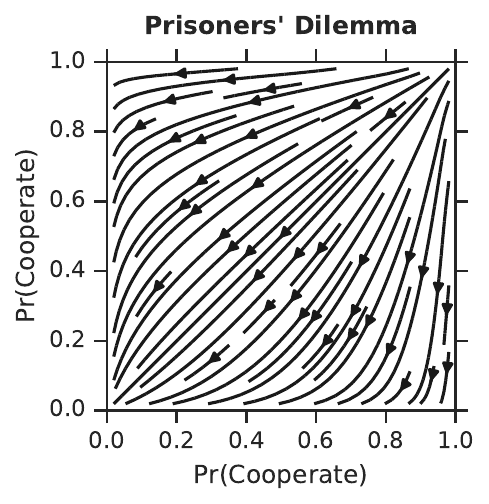} & 
      \includegraphics[scale=0.45]{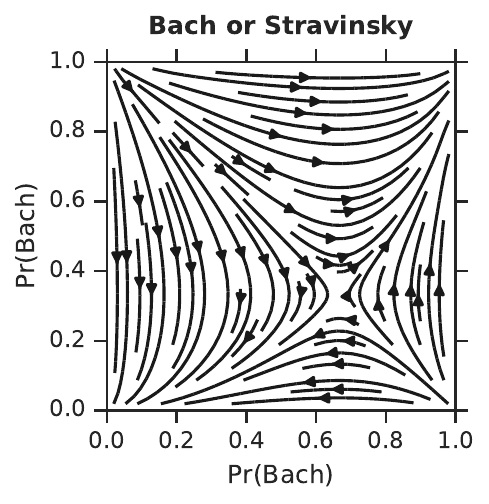} &
      \includegraphics[scale=0.45]{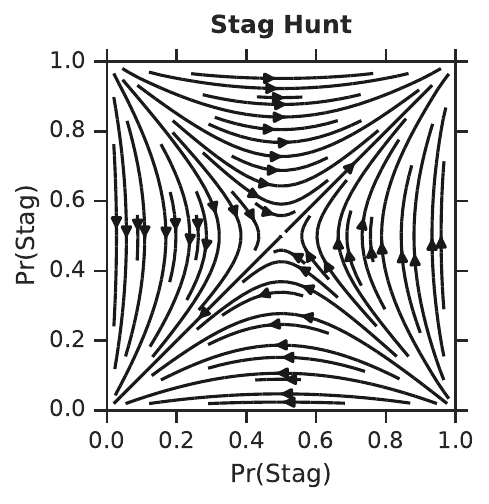} \\      \includegraphics[scale=0.45]{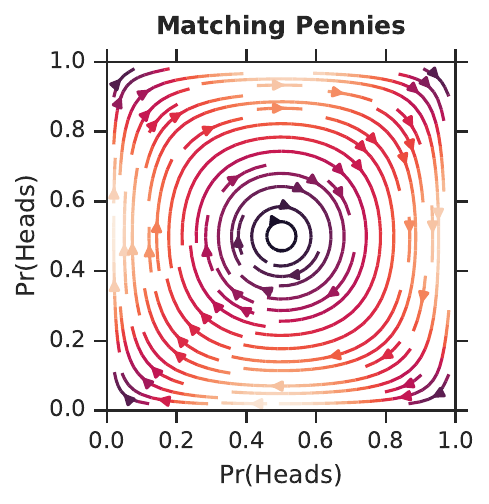} &
      \includegraphics[scale=0.45]{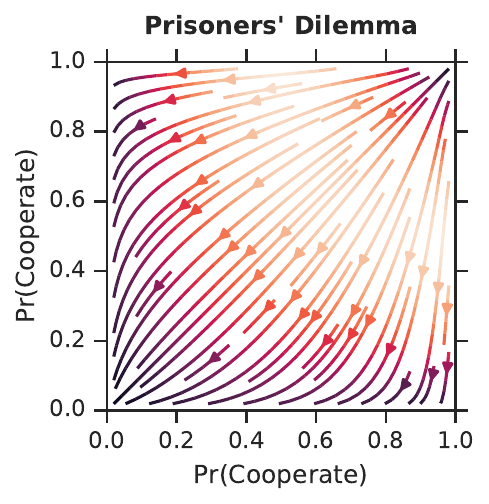} &
      \includegraphics[scale=0.45]{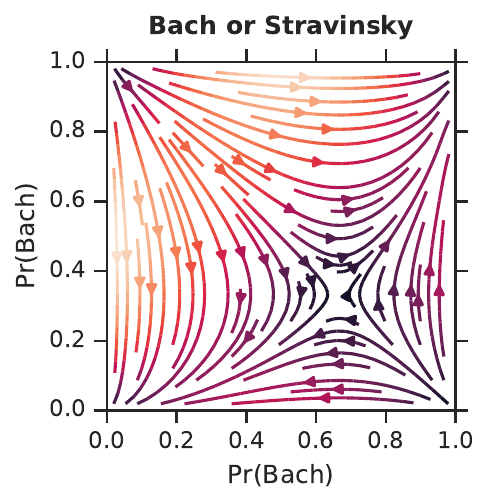} &
      \includegraphics[scale=0.45]{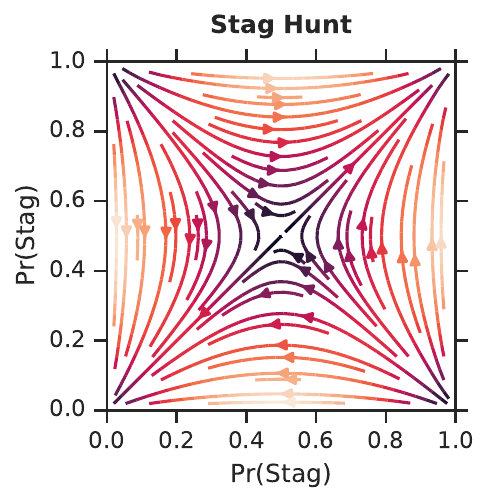} \\
    \end{tabular}
    \caption{Phase portraits of the two-population replicator dynamics for four common
    bimatrix games. The colored plots shows the relative magnitude of the vectors.}
    \label{fig:rd-dynamics}
\end{figure}

Figure~\ref{fig:rd-dynamics-rps} shows plots generated from OpenSpiel for replicator dynamics in the game of Rock--Paper--Scissors.
Figure~\ref{fig:rd-dynamics} shows plots generated from OpenSpiel for four common bimatrix games.


\subsubsection{$\alpha$-Rank}

$\alpha$-Rank \cite{Omidshafiei19AlphaRank} is an algorithm that leverages evolutionary game theory to rank AI agents interacting in multiplayer games. Specifically, $\alpha$-Rank defines a Markov transition matrix with states corresponding to the profile of agents being used by the players (i.e., tuples of AI agents), and transitions informed by a specific evolutionary model that ensures correspondence of the rankings to a game-theoretic solution concept known as a Markov-Conley Chain. A key benefit of $\alpha$-Rank is that it can rank agents in scenarios involving intransitive agent relations (e.g., the agents Rock, Paper, and Scissors in the eponymous game), unlike the Elo rating system \cite{Balduzzi18Evaluation}; an additional practical benefit is that it is also tractable to compute in general games, unlike ranking systems relying on Nash equilibria \cite{daskalakis2009complexity}.

OpenSpiel currently supports using $\alpha$-Rank~ for both single-population (symmetric) and multi-population games. Specifically, users may specify
games via payoff tables (or tensors for the >2 players case) as well as Heuristic Payoff Tables (HPTs).
Note that here we only include an overiew of the technique and visualizations; for a tour through the usage and code please see
the \href{https://github.com/deepmind/open_spiel/blob/master/docs/alpha_rank.md}{$\alpha$-Rank doc on the web site}.

\begin{figure}[h!]
    \centering
    \begin{tabular}{cc}
    \vspace{1cm}{\color{white}.} & \multirow{3}{*}{\includegraphics[scale=0.4]{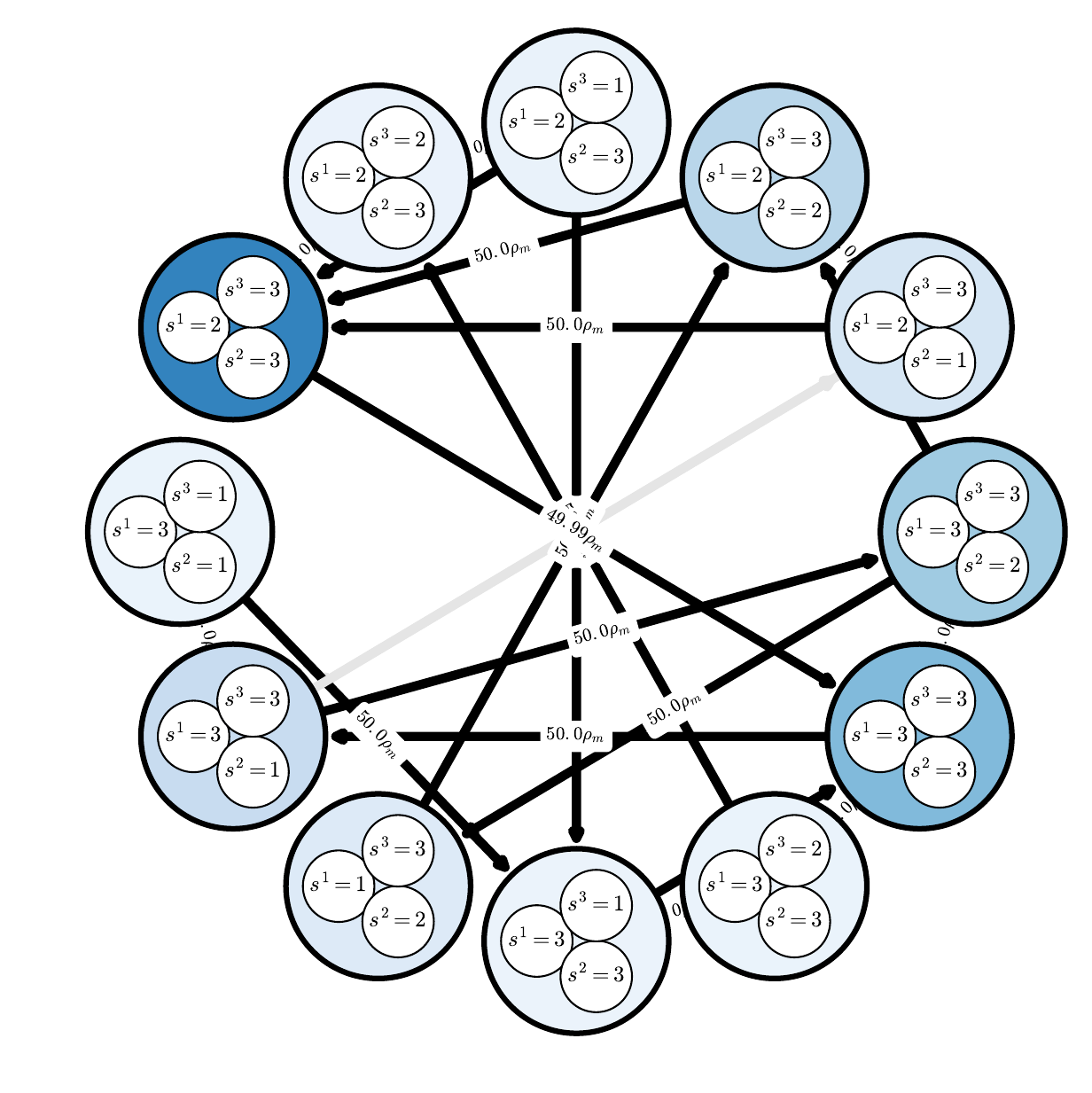}} \\
    \includegraphics[scale=0.33]{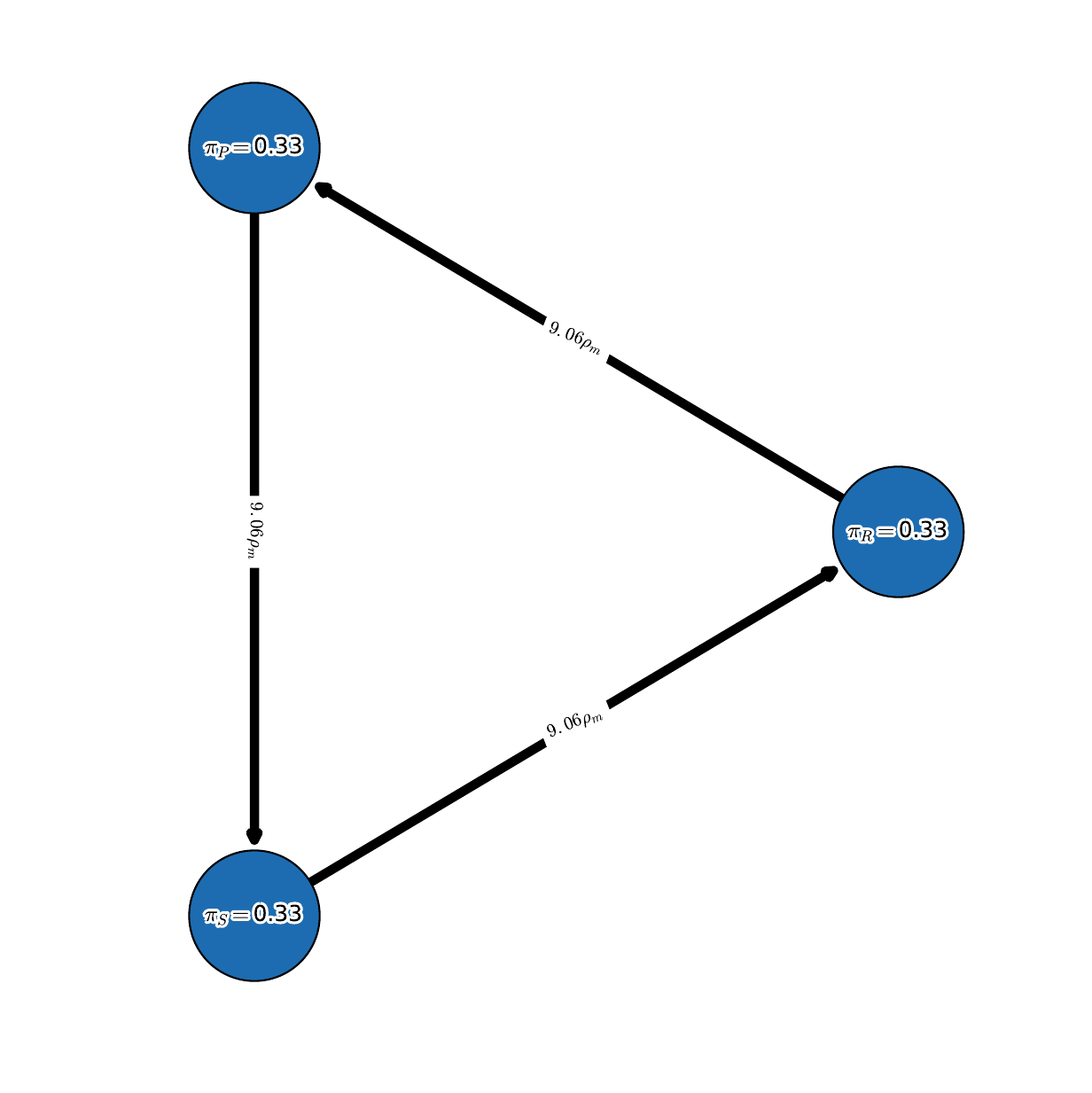} &
    \vspace{0.5cm}{\color{white}.} \\
    (a) & (b)\\
    \end{tabular}
    \caption{(a) Markov transitions matrix of solution found by $\alpha$-Rank on Rock, Paper, Scissors. (b) Markov transitions matrix of meta-game computed by the first few rounds of XFP in 3-player Kuhn poker.}
    \label{fig:alpha-rank-mcc}
\end{figure}

Figure~\ref{fig:alpha-rank-mcc}(a) shows a visualization of the Markov transition matrix of $\alpha$-Rank run on
the Rock, Paper, Scissors game.
The next example demonstrates computing $\alpha$-Rank on an asymmetric 3-player meta-game, constructed by computing utilities for Kuhn poker agents from the
best response policies generated in the first few rounds of via extensive-form fictitious play (XFP)~\cite{Heinrich15FSP}. 
The result is shown in Figure~\ref{fig:alpha-rank-mcc}(b).

\begin{figure}[h!]
    \centering
    \includegraphics[scale=0.45]{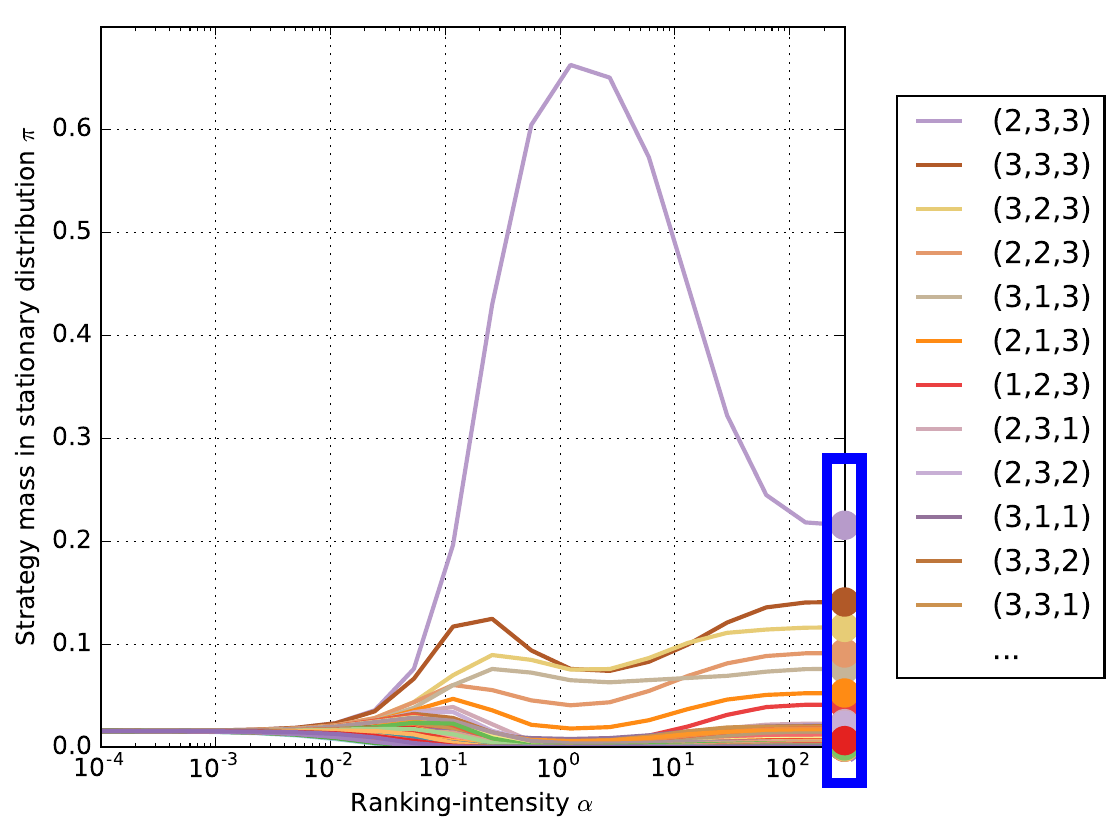}
    \caption{Effect of ranking-intensity parameter $\alpha$ on policy mass in stationary distribution in meta-game generated by XFP in 3-player Kuhn poker.}
    \label{fig:alpha-rank-3pkuhn-pi-vs-alpha}
\end{figure}

One may choose to conduct a sweep over the ranking-intensity parameter, $\alpha$ (as opposed to choosing a fixed $\alpha$). This is useful for general games where bounds on utilities may be unknown, and where the ranking computed by $\alpha$-Rank should use a sufficiently high value of $\alpha$ (to
ensure correspondence to the underlying Markov-Conley Chain solution concept). In such cases, the following interface can be used to both visualize the
sweep and obtain the final rankings computed. The result is shown in Figure~\ref{fig:alpha-rank-3pkuhn-pi-vs-alpha}.

\section{Guide to Contributing}

If you are looking for ideas on potential contributions or want to see a rough road map for the future of
OpenSpiel, please visit the
\href{https://github.com/deepmind/open_spiel/blob/master/docs/contributing.md}{Roadmap and Call for Contributions on
github}.

Before making a contribution to OpenSpiel, please read the design philosophy in Section~\ref{sec:design-api}.
We also kindly request that you contact us before writing any large piece of code, in case (a) we are already
working on it and/or (b) it's something we have already considered and may have some design advice on its
implementation. Please also note that some games may have copyrights which could require legal approval(s).
Otherwise, happy hacking!

\subsection{Contacting Us}

If you would like to contact us regarding anything related to OpenSpiel, please create an issue on the 
\href{https://github.com/deepmind/open_spiel}{github site} so that the team is notified, and
so that the responses are visible to everyone.

\bibliography{main}
\bibliographystyle{plain}

\end{document}